\documentclass[sigconf]{acmart}

\usepackage{graphicx}
\usepackage{microtype}
\usepackage{hyperref}
\usepackage[capitalise]{cleveref}
\usepackage{booktabs}
\usepackage{tabulary}
\usepackage[british]{babel}
\usepackage{csquotes}
\usepackage{makecell}
\usepackage[e]{esvect}
\usepackage{bm}
\usepackage[nolist]{acronym}
\usepackage[separate-uncertainty=true, detect-weight=true, detect-family=true]{siunitx}
\usepackage{multirow}

\setcopyright{none}
\renewcommand\footnotetextcopyrightpermission[1]{} 

\newcommand{\vc}[1]{\multirow{1.75}{*}{#1}}

\begin{acronym}[SRAM]
	\acro{nn}[NN]{neural network}
	\acro{agn}[AGN]{additive Gaussian noise}
	\acro{sram}[SRAM]{Static Random Access Memory}
	\acro{mc}[MC]{Monte Carlo}
	\acro{sgd}[SGD]{Stochastic Gradient Descent}
	\acro{ilp}[ILP]{Integer Linear Programming}
	\acro{qat}[QAT]{Quantization-aware Training}
	\acro{ste}[STE]{Straight-Through Estimator}
	\acro{mre}[MRE]{Mean Relative Error}
	\acro{am}[AM]{approximate multiplier}
	\acro{cnn}[CNN]{convolutional neural network}
	\acro{fc}[FC]{Fully-Connected}
	\acro{pstl}[PSTL]{parametric signal temporal logic}
	\acro{mac}[MAC]{Multiply-Accumulate Operation}
	\acro{qos}[QoS]{Quality of Service}
	\acro{mse}[MSE]{mean squared error}
\end{acronym}

\begin{document}
	
	\title{QoS-Nets: Adaptive Approximate Neural Network Inference}
	
	\author{Elias Trommer}
	\email{elias.trommer@infineon.com}
	\affiliation{%
		\institution{Infineon Technologies}
		\streetaddress{Königsbrücker Straße 180}
		\postcode{01099}
		\city{Dresden}
		\country{Germany}
	}
	\author{Bernd Waschneck}
	\email{bernd.waschneck@infineon.com}
	\affiliation{%
		\institution{Infineon Technologies}
		\streetaddress{Königsbrücker Straße 180}
		\postcode{01099}
		\city{Dresden}
		\country{Germany}
	}
	\author{Akash Kumar}
	\email{akash.kumar@tu-dresden.de}
	\affiliation{%
		\institution{Center for Advancing Electronics Dresden (cfaed)}
		\city{Dresden}
		\country{Germany}
	}
	
	\begin{abstract}
		In order to vary the arithmetic resource consumption of neural network applications at runtime, this work proposes the flexible reuse of approximate multipliers for neural network layer computations. We introduce a search algorithm that chooses an appropriate subset of approximate multipliers of a user-defined size from a larger search space and enables retraining to maximize task performance. Unlike previous work, our approach can output more than a single, static assignment of approximate multiplier instances to layers. These different operating points allow a system to gradually adapt its \ac{qos} to changing environmental conditions by increasing or decreasing its accuracy and resource consumption. QoS-Nets achieves this by reassigning the selected approximate multiplier instances to layers at runtime. To combine multiple operating points with the use of retraining, we propose a fine-tuning scheme that shares the majority of parameters between operating points, with only a small amount of additional parameters required per operating point. In our evaluation on MobileNetV2, QoS-Nets is used to select four approximate multiplier instances for three different operating points. These operating points result in power savings for multiplications between 15.3\% and 42.8\% at a Top-5 accuracy loss between 0.3 and 2.33 percentage points. Through our fine-tuning scheme, all three operating points only increase the model's parameter count by only 2.75\%.
	\end{abstract}
	
	
	\maketitle
	\pagestyle{plain} 
	\acresetall 
	
	\section{Introduction}
	The recent growth in the capabilities of artificial~\acp{nn} has been enabled by an explosion of parameter counts and the amount of compute required for training and inference~\cite{sevillaComputeTrendsThree2022}. To keep up with the rapidly increasing complexity of \ac{nn}'s, several inference optimization techniques exist. All of these techniques aim to make models run more efficiently while maintaining high task performance. Pruning and Quantization~\cite{hanDeepCompressionCompressing2016}, two common examples of \ac{nn} optimizations, target the complexity of \ac{nn}'s at the operand level. Another set of techniques, broadly referred to as \enquote{Approximate Computing}, instead addresses the complexity of the \emph{operator} itself: A major driver of inference cost for \ac{nn}'s is the matrix multiplication, which in turn can be broken down further into \acp{mac}. Here, particularly the multiplication is costly to implement. Approximate implementations of hardware multiplication units trade strict mathematical accuracy under all circumstances for a reduction in the consumption of other resources.
	
	An extensive body of work discusses how to best apply these \acp{am} in the context of \acp{nn}~\cite{zervakisApproximateComputingML2021}. An important benefit of \acp{am} is that, for the same sets of operands, they can provide a nearly arbitrary number of operators with varying levels of accuracy. This allows for flexible and highly customizable solutions. To provide the best compromise of task performance and resource consumption, recent work has shown that two components are vital: First, retraining the network's parameters while simulating the inaccurate multiplication is crucial to minimizing the propagated error that occurs whenever \ac{am} results are used as inputs to subsequent \ac{nn} layers~\cite{delaparraFullApproximationDeep2020}. Secondly, not all layers in a network are equally sensitive to inaccurate computation. Looking at each layer individually when deciding the tolerable level of inaccuracy has been shown to provide much better performance~\cite{trommerCombiningGradientsProbabilities2022}, compared to a one-fits-all approach where the same \ac{am} is assigned to all layers.
	
	The second point in particular has led to the emergence of different hardware paradigms which aim to enable flexible approximate products. Tile-based accelerators~\cite{mrazekALWANNAutomaticLayerWise2019} provide a fixed set of compute tiles dedicated to matrix multiplications. Each tile uses a different \ac{am} for product computation. The optimization problem then consists of selecting an appropriate \ac{am} instance for each tile and then mapping the \ac{nn}'s layers to the different tiles in a way that simultaneously maximizes task accuracy and minimizes resource consumption. Alternatively, several \acp{am} might be merged into a single hardware unit which allows the approximate product to be configured at runtime~\cite{tasoulasWeightOrientedApproximationEnergyEfficient2020, spantidiPositiveNegativeApproximate2021}. This work does not focus on any specific hardware implementation, but develops an optimization algorithm that is meant to be broadly applicable.
		
	Although a substantial amount of literature discusses certain aspects of this optimization there is, to date, no publication that unifies these building blocks in order to provide an answer to the questions: Given a large set of \acp{am} and a \ac{nn}, which subset of $n$ \acp{am} should be chosen to provide the best compromise of accuracy and resource consumption? How can these \acp{am} be mapped to individual \ac{nn} layers without testing a large number of candidate solutions, so that the expensive but important retraining step becomes feasible?
	
	Existing literature thus far only derives a single, static configuration for a network. This does not fully exploit the flexibility of the many different possible \ac{am} design points and hardware platforms with selectable levels of approximation. Instead of one static configuration, we propose to reuse a platform's set of \acp{am} in more than one way: With seamless switching between different levels of operator accuracy, our method can provide several \emph{operating points}. Each operating point assigns the \acp{am} available in the system differently to the \ac{nn}'s layers, in order to strike a different balance of task performance and resource consumption. The method does not rely on any specific hardware architecture but might be implemented using any platform that supports runtime adaptation of the multiplication operator. Depending on environmental conditions, a platform can then choose to provide higher task performance at the cost of increased resource consumption, or reduced accuracy with lower resource consumption. Having more than one operating point allows for gradually adjusting the platform's \ac{qos} through switching from one operating point to another.
	The novel contributions of this work are:
	\begin{itemize}
		\item  An optimization algorithm that combines gradient-based optimization with Clustering to select a fixed number of \acp{am} from a larger search space and allows for retraining. A guaranteed number of output \ac{am} instances allows the user to limit the number of arithmetic units that a platform would need to implement. 
		\item A method that allows for the search procedure to produce not just a single, but multiple Pareto points with different trade-offs between task performance and resource consumption: the network's \emph{operating points}. Our method optimizes the choice of \ac{am} simultaneously across all operating points to provide a balanced set of \acp{am}.
		\item A parameter-efficient retraining method that allows for the majority of the network's parameters to be shared by all its operating points.
	\end{itemize}
	\section{Background and Related Work}
	Previous works have studied a wide array of algorithms to address the mapping problem of deciding which portion of a network is run using which approximate operator. An important work in this field is ALWANN~\cite{mrazekALWANNAutomaticLayerWise2019}. The authors propose an accelerator architecture that consists of $n$ compute tiles. Each tile is implemented using a different \ac{am} which is selected from EvoApproxLib~\cite{mrazekLibrariesApproximateCircuits2020}, a library with a large number of \acp{am}. A genetic algorithm simultaneously determines a subset $n$ \acp{am} from the search space that should be used and decides which of the tiles will execute which \ac{nn} layer. Instead of retraining, a weight-tuning scheme partially compensates for the error introduced by the inaccurate compute tiles. Spantidi et al.~\cite{spantidiPerfectMatchSelecting2023} discuss a similar problem: \ac{pstl} is used to first choose a subset of $n$ \acp{am} which are merged into a single hardware component that can be reconfigured at runtime. Subsequent \ac{pstl} queries then determine which range of weight values in each layer will be mapped to which level of approximation. Other works do not choose from a larger set of \acp{am}, but start out with a fixed \ac{am} design that is reconfigurable at runtime. LVRM~\cite{tasoulasWeightOrientedApproximationEnergyEfficient2020} uses a divide and conquer approach to map value ranges to different \ac{am} modes. PNAM~\cite{spantidiPositiveNegativeApproximate2021} relies on a specific \ac{am} design which allows for mapping value ranges in a way that minimizes the accumulated error. All these works boost the achievable accuracy through mapping decisions or heuristics that tune pre-trained weights. De la Parra et al.~\cite{delaparraFullApproximationDeep2020} first demonstrated that retraining network parameters to recover accuracy is highly beneficial. The work, however, only considered configurations in which a single \ac{am} is deployed throughout the entire network. Trommer et al.~\cite{trommerCombiningGradientsProbabilities2022} then combined retraining with heterogeneous \ac{am} configurations. The authors use \ac{agn} as a proxy for each layer's sensitivity to approximation errors. The noise injected into each layer is optimized using gradient descent. An error model is proposed, which converts each \ac{am}'s error function as well as layerwise parameter and input distribution samples into an estimate of error mean and standard deviation. Based on this estimate, an \ac{am} is chosen for each layer. The work does, however, lack a mechanism to constrain the result to a subset of the $n$ most useful \acp{am} from the search space. This limitation might make it less practical for real world applications where layers often share compute hardware. We provide an overview of these different methods in~\Cref{tab:literature}.
	\begin{table}[tb]
		\caption{Comparison of mapping algorithms for operator-based approximations to neural network parameters}
		\footnotesize
		\begin{tabulary}{\linewidth}{LCCCC}
			\toprule
			Publication & Constrained Choice & Retraining & Algorithm & Granularity \\
			\midrule
			TPM~\cite{spantidiPositiveNegativeApproximate2021} & $\checkmark$ & --- & PSTL & Value \\
			ALWANN~\cite{mrazekALWANNAutomaticLayerWise2019} & $\checkmark$ & --- & Genetic & Layer \\
			LVRM~\cite{tasoulasWeightOrientedApproximationEnergyEfficient2020} & --- & --- & D\&C & Value \\
			Gradient Search~\cite{trommerCombiningGradientsProbabilities2022} & \vc{---} & \vc{$\checkmark$} & \vc{Gradient} & \vc{Layer} \\
			QoS-Nets (this work) & $\checkmark$ & $\checkmark$ & Gradient + Clustering & Layer \\
			\bottomrule
		\end{tabulary}
		\label{tab:literature}
	\end{table}

	
	\section{Proposed Methodology}
	We extend an unconstrained heterogeneous mapping algorithm from the literature to allow for constraining the number of \ac{am} instances in the solution to a fixed amount by applying k-Means Clustering~\cite{lloydLeastSquaresQuantization1982} to intermediate results. This is followed by a method for deriving several operating points from a heterogeneous mapping solution. The proposed clustering algorithm is extended to simultaneously optimize the choice of \acp{am} across \emph{all} operating points. A fine-tuning procedure that facilitates the sharing of parameters across several operating points is proposed in order to maximize task performance while minimizing memory overhead.
	
	\subsection{$n$-constrained Multiplier Selection}
	\label{sec:selection}

\begin{figure}[tb]
		\centering
		\includegraphics[width=\linewidth]{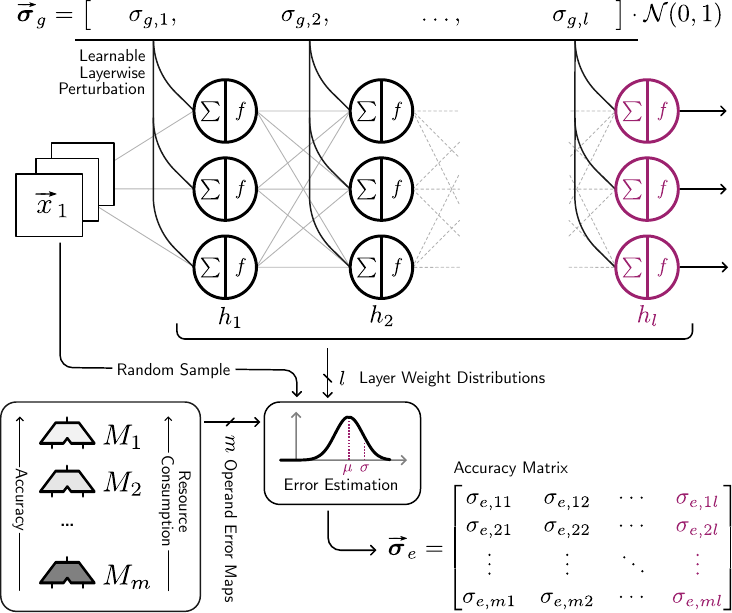}
		\caption{Preparation of an $l \times m$ error estimation matrix as described in~\cite{trommerCombiningGradientsProbabilities2022}}
		\label{fig:search_space}
	\end{figure}

	The proposed constrained selection approach improves upon the algorithm that was put forward by Trommer et al.~\cite{trommerCombiningGradientsProbabilities2022}. In its original form, the method perturbs the computation of the \ac{nn}'s output by adding a variable amount of \ac{agn} to each layer. The amount of noise for a network with $l$ layers can be described as a vector of positive, real numbers that scale the amount of random noise which gets injected into each layer's output during training $\vv{\bm{\sigma}}_g \cdot \mathcal{N}(0,1), \; \vv{\bm{\sigma}}_g \in \mathbb{R}_+^l$. The vector $\vv{\bm{\sigma}}_g$ is differentiable with respect to the model's training loss, so it can be optimized using gradient descent. The authors extend the loss function to avoid situations where the loss would be entirely dominated by the injected noise or the task loss. To relate the optimized amount of injected noise to actual \ac{am} hardware, the authors then propose an error model which predicts each \ac{am}'s standard deviation for each layer, based on the layer's weights and activations sampled from the training data. For $m$ \acp{am} of interest, this yields an $l \times m$ matrix $\vv{\bm{\sigma}}_e \in \mathbb{R}_+^{l,m}$ with estimates of the error standard deviation for each combination of layer parameters and \ac{am}. The process is depicted in~\Cref{fig:search_space}. The error mean is ignored in this method, since it can be compensated for when the network is retrained. For each layer $k$, only those \ac{am} instances $j$ with $\sigma_{e,jk} \le \sigma_{g,k}$ are considered accurate enough. From this remaining set, the instance that minimizes another metric like power consumption is selected in the original work.

	The method does not exchange information across layers. This makes it impractical, since the chosen hardware instances might not be reusable throughout the network. Unconstrained choice means that any solution might contain up to $\min (m, l)$ \acp{am} at worst. We improve upon this by constraining the search procedure so that it produces a heterogeneous mapping that uses at most $n$ instances from the search space, with $n$ being predetermined by the user according to the target system's properties.
	
	To limit the solution space to a fixed number of \acp{am}, we apply the k-Means Clustering algorithm~\cite{lloydLeastSquaresQuantization1982} to the intermediate results of the original algorithm. The aim is to find groups of layers that have similar accuracy requirements and let them share \ac{am} instances instead of directly assigning the best choice for each layer. To create the input space for the clustering, we first discard the entries in $\vv{\bm{\sigma}}_e$ that do not deliver usable accuracy on any layers, i.e. the rows corresponding to \ac{am} instances $j$ where $\vv{\bm{\sigma}}_{e,jk} \ge \vv{\bm{\sigma}}_{g,k};\; 0 < k \le l$, since these will not be part of the solution set and can safely be excluded. Next, we split the columns of $\vv{\bm{\sigma}}_e$ into one preference vector $\vv{\sigma}_{bk}$ for each layer $0 < k \le l$. 
	\begin{align}
	\vv{\sigma}_{bk} := \frac{\vv{\bm{\sigma}}_{e}(:,k)}{\sigma_{g,k}}
	\end{align}
	Intuitively, each preference vector expresses the accuracy requirement of the layer it corresponds to. Every entry in the preference vector refers to one \ac{am} instance's error standard deviation with respect to the layer's task. The preference vectors are normalized to the corresponding layer's required accuracy as expressed in $\vv{\bm{\sigma}}_g$ to make them comparable across layers. Entries with a value of more than one refer to \acp{am} with insufficient accuracy for the layer, those with values below one to instances that meet the layer's accuracy preference. Our clustering space $\mathcal{C}$ then becomes:
	\begin{align}
		\mathcal{C} = \left\lbrace \vv{\sigma}_{b1}, \vv{\sigma}_{b2}, \dots, \vv{\sigma}_{bl} \right\rbrace
	\end{align}
 	We notice that some entries in the preference vectors are very high, meaning that the corresponding \ac{am} instance is well below the required accuracy for the given layer. k-Means Clustering minimizes the sum of squared distances in each cluster. If left unchecked, these outlier values would have a very strong influence on the results, despite them likely being irrelevant to the actual solution. However, they can also not simply be replaced with a fixed value or omitted entirely. Because we try to find clusters of layers with similar error requirements, the accuracy requirement for each layer in a cluster becomes a soft, rather than a hard constraint. Simply discarding instances that do not meet the accuracy required by a single layer might exclude instances that would be a reasonable compromise for \emph{all} layers in the cluster. Instead, we rescale those \ac{am} instances that don't reach the required accuracy for a layer in order to reduce their impact without losing their important relative ordering. Applying a monotonically increasing transformation that pulls high values closer to a value of one retains the relative ordering of \ac{am} instances and thus the information about their performance, but limits their drag on the clustering result. We apply a continuous and monotonically increasing reweighting function
	\begin{align}
		f\left(x \right) = \begin{cases}
			x \,&\text{if}\,x \le 1 \\
			1+\ln\left(x\right) &\text{otherwise}
		\end{cases}
	\end{align}
	to each element in every preference vector $\vv{\sigma}_{ok} \in \mathcal{C}$. $f$ moves outliers with large magnitudes closer to one, leading to lower impact on the clustering while retaining all the relevant information.
	
	On this transformed input space, k-Means Clustering is applied to build clusters of preference vectors. The amount of desired clusters is set to $n$, the number of \ac{am} instances that a solution is allowed to consist of. We then obtain $n$ centroids and $l$ cluster assignments. The centroid of each cluster has the same dimensionality and structure as the original preference vectors $\vv{\bm{\sigma}}_{bk}$. Every entry in a centroid refers to the average performance of an \ac{am} instance across all layers that are assigned to the cluster. This representation makes it easy to pick an \ac{am} instance for each cluster: Of all entries in the centroid, those with a normalized error standard deviation of less than one provide sufficient accuracy for the layers in the cluster. From those remaining instances, the one that minimizes resource consumption is picked and reused for all layers assigned to the cluster.
	
	\subsection{Multiple approximate operating points}
	To increase the flexibility of the resulting configuration further, we extend the clustering approach from~\Cref{sec:selection} in a way that allows the output to reuse the subset of \acp{am} that are chosen in more than one way. For $o=1$ operating points, each layer is assigned an \ac{am} instance which remains fixed throughout the lifetime of the application. With multiple operating points $o > 1$, the system might switch between inference modes with higher or lower accuracy, depending on environmental conditions. When high accuracy is required, a mode with more accurate \ac{am} instances can be chosen. When resources are scarce, the system has the option to switch to a mode with lower accuracy, gracefully reducing the \ac{qos}. Rather than using clustering to find one operating point and then deriving others using a heuristic, we would like the search procedure to consider all operating points as equally important and choose the multiplier subset so that it provides a good compromise of \ac{am} instances across the different operating points. We first define one scale factor for each operating point $\mathcal{S} := \left\lbrace s_1, s_2, \dots, s_o \right\rbrace$. The clustering input space is then expanded by rescaled versions of each layer's preference vectors.
	\begin{align}
		\mathcal{C}' = \left\lbrace s \cdot \vv{\sigma}_b \mid s \in \mathcal{S}, \vv{\sigma}_b \in \mathcal{C} \right\rbrace
	\end{align}
	\begin{figure}[tb]
		\centering
		\includegraphics[width=\linewidth]{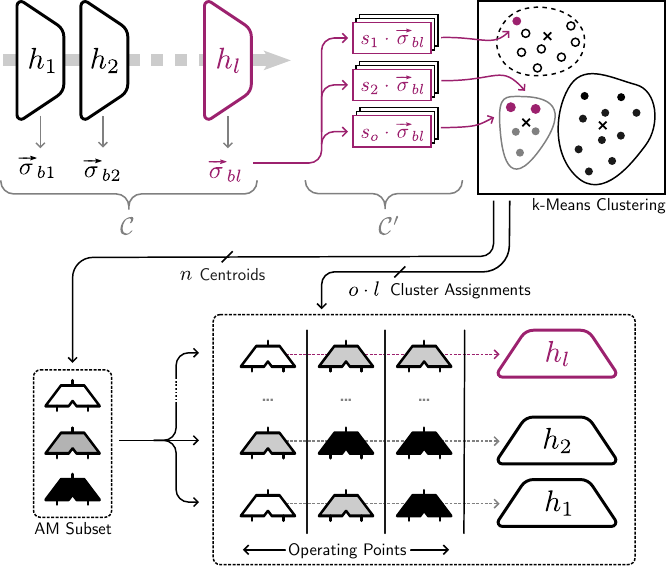}
		\caption{Scaling and Clustering of preference vectors to find \ac{am} subset and assignment of \ac{am} instance to layers for different operating points}
		\label{fig:clustering}
	\end{figure}
	Every preference vector in our clustering space now corresponds to a combination of layer and operating point. Lower values for $s$ correspond to higher accuracy, while higher values of $s$ minimize resource consumption at the cost of degraded accuracy. As before, we run k-Means Clustering on this extended search space and obtain $n$ centroids and $o \cdot l$ cluster assignments as shown in~\Cref{fig:clustering}. From each centroid, an \ac{am} instance can be chosen in exactly the same way as was discussed before. This \ac{am} instance is then reused for all layer/operating point combinations that are assigned to its cluster.
	\subsection{Low-Overhead Fine-Tuning}
	The effort required for dynamically reconfiguring the multiplication operator is determined by the  hardware implementation used by the target platform. For a tile-based accelerator with compute tiles of varying precision~\cite{mrazekALWANNAutomaticLayerWise2019}, changing the routing of intermediate results is required when switching between operating points. For hardware where the multiplier's logic is reconfigurable using multiplexers, a switching from one logic path to another is required to change operating points. In both cases, we assume that the resource consumption of the reconfiguration will be dominated by the resource consumption of the inference itself. The same logic does, however, not hold for the network's parameters. Previous work has demonstrated that retraining of \ac{nn} parameters is highly beneficial for compensating the systematic error introduced by \acp{am}~\cite{delaparraFullApproximationDeep2020}. When introducing several operating points, a naive retraining approach would add significant memory requirements, because every operating point would require its own full set of weights and biases. Switching from one operating point to another would require loading a completely new set of \ac{nn} parameters, potentially stressing the system's memory interface. Depending on an implementation's architecture, this increased memory traffic and consumption is likely detrimental to overall performance. Conceptually, switching between operating points should be a lightweight operation with minimal additional cost. To avoid the penalty of full retraining for each operating point, we propose the use of a fine-tuning strategy instead of full retraining of all parameters. By focusing on fine-tuning a select set of \ac{nn} parameters, we aim to materialize most of the gains from retraining, although with a much lower parameter overhead. 
	
	An important concept in the context of \acp{am} is that of \emph{covariate shift}~\cite{ioffeBatchNormalizationAccelerating2015}. Assume that an \ac{am} instance adds an error $X$ to a layer's output. If $\mathbb{E}[X] \ne 0$, then the \ac{am} instance exhibits a \emph{systematic} error, which causes a shift in the output distribution of the layer's results. Because a deep learning model is a chained computation over a stack of layers, this error propagates throughout the model and is even exacerbated if the same \ac{am} instance is used throughout multiple layers, which explains why deeper models tend to suffer a faster drop in performance when deploying \acp{am}~\cite{tasoulasWeightOrientedApproximationEnergyEfficient2020}. To understand this behavior better, we can investigate the \ac{mse} of $X$ with respect to the layer's output under accurate computation. It can be expressed using the equality $\mathbb{E}(X^2) = \operatorname{Var}(X) + \mathbb{E}[X]^2$. The second term $\mathbb{E}[X]^2$ corresponds to the \ac{am}'s systematic error and has a quadratic impact on the \ac{mse}. However, this systematic error can be compensated for by adjusting each neuron's bias $b$ to minimize the propagated approximation error mean $\mathbb{E}[X]$ so that $b' = b - \mathbb{E}[X]$. In practice, the exact value of $\mathbb{E}[X]$ is not known in advance, since it is determined by the values of each neuron's activations and weights. The minimization is therefore best carried out by the \ac{nn} training process itself.

	Based on this reasoning, we propose to adjust only the network's biases while keeping the weights frozen. Minimizing the approximation error's mean is likely sufficient to restore most of the network's accuracy. Biases only contribute a small amount to the network's total memory footprint, so having separate sets of biases per operating points will have little impact on the application's memory consumption. During training, the task of compensating systematic errors in intermediate results is taken over by BatchNormalization layers~\cite{ioffeBatchNormalizationAccelerating2015}. To provide operating points with few additional parameters, we freeze the network's weights during retraining and only optimize each layer's BatchNormalization parameters $\gamma_l$ and $\beta_l$. For inference, these separate parameters can efficiently be combined with the layer's bias addition as well as integer quantization and requantization steps.
	 
	\section{Results and Discussion}
	All experiments were carried out on a host system equipped with an AMD Ryzen 9 3900X CPU and a single nVidia RTX 2080Ti GPU. We implement our search algorithm as an extension of the popular deep learning framework PyTorch~\cite{paszkePyTorchImperativeStyle2019} and use the open-source library TorchApprox~\cite{trommerHighThroughputApproximateMultiplication2023} for simulation of approximate operators and retraining. 8-bit quantization is the default numerical format for all models and the search space is formed by the 37 8\texttimes8-bit unsigned \acp{am} from the most recent version of the EvoApprox library~\cite{mrazekLibrariesApproximateCircuits2020} at the time of writing. In line with the previous work used for comparison, relative power consumption is calculated as a weighted sum of the power consumption of each layer's assigned \ac{am} instance, scaled by the amount of multiplications in the respective layer. The power consumption of each \ac{am} is expressed relative to that of the accurate EvoApprox instance after synthesis to \texttt{PDK45}. \ac{sgd} with a momentum of $0.9$ is used for all optimization runs.
	
	\subsection{Constrained Multiplier Choice}
	We compare the performance of our clustered mapping algorithm to other works from the literature that address similar problems. To the best of our knowledge, no previous work derives multiple operating points from a single set of \acp{am}. This means that all search algorithms in our comparison optimize the choice of \acp{am} for $o=1$, i.e. a single, static operating point only. We optimize the constrained \ac{am} choice for several sizes of the ResNet architecture~\cite{heDeepResidualLearning2016} on the CIFAR-10 and CIFAR-100~\cite{krizhevskyLearningMultipleLayers2009} datasets. The EvoApprox~\cite{mrazekLibrariesApproximateCircuits2020} library of Approximate Arithmetic Circuits is used as the search space of approximate multiplication operators. This is a commonly-used experimental setup which allows for a direct comparison.
	
	Each ResNet variant is trained using the hyperparameters and augmentations described in the original work, followed by 30 epochs of \ac{qat}. We then optimize the noise injection for another 10 epochs. This determines each layer's sensitivity to small perturbation (its accuracy preference). With this information, we apply k-Means Clustering to group the preference vectors of each layer into $n$ clusters. An \ac{am} instance is chosen for each cluster centroid and deployed to the respective layers. The resulting configuration is retrained for 10 epochs with a learning rate of $1\cdot10^{-3}$ and a learning rate decay of 90\% after five and eight epochs respectively.
	
	We compare to ALWANN~\cite{mrazekALWANNAutomaticLayerWise2019}, which assumes a tile-based accelerator with a choice of up to 4 different compute tiles. PNAM~\cite{spantidiPerfectMatchSelecting2023} and LVRM~\cite{tasoulasWeightOrientedApproximationEnergyEfficient2020} both propose a single, reconfigurable \ac{am} design with three different approximation modes each. Because the output space is predetermined, this is not strictly a constrained choice problem. The work, however, targets a similar application, which is why we include a comparison with our method for $n=3$ in these cases as well. TPM~\cite{spantidiPerfectMatchSelecting2023} discusses optimizing the choice of a useful subset from the EvoApprox library for the more complex CIFAR100 dataset, albeit for a deployment without retraining. We further include a homogeneous solution (i.e. one that uses a single \ac{am} instance throughout the entire network), retrained using the same hyperparameters as our heterogeneous solution.
	
		\begin{table}[tb]
		\caption{Comparison of power consumption reduction and accuracy loss on CIFAR-10 for different ResNet variants and multiplier selection methods}
		\footnotesize
		\begin{tabulary}{\linewidth}{llCC}
			\toprule
			Model & Method & Power Reduction \textuparrow & Top-1 Accuracy Loss [p.p.] \textdownarrow \\
			\midrule
			ResNet8		& ALWANN~\cite{mrazekALWANNAutomaticLayerWise2019}								& \qty{30}{\percent}		& 1.7 \\
			{}			& Homogeneous~\cite{delaparraFullApproximationDeep2020}, \texttt{mul8u\_179B}	& \textbf{47 \%} 		& 1.5 \\
			{}			& QoS-Nets, $o=1$, $n=4$														& \qty{41}{\percent}		& \textbf{0.8} \\
			\midrule
			ResNet14	& ALWANN~\cite{mrazekALWANNAutomaticLayerWise2019}								& \qty{30}{\percent}		& 0.9 \\
			{}			& Homogeneous~\cite{delaparraFullApproximationDeep2020}, \texttt{mul8u\_179B}	& \textbf{47 \%} 		& 0.9 \\
			{}			& QoS-Nets, $o=1$, $n=4$														& \qty{46}{\percent}		& \textbf{0.8} \\
			\midrule
			ResNet20	& LVRM~\cite{tasoulasWeightOrientedApproximationEnergyEfficient2020}			& \qty{17}{\percent}		& 0.5 \\
			{} 			& PNAM~\cite{spantidiPositiveNegativeApproximate2021}
			& \qty{19}{\percent}	& 0.5 \\
			{}			&  Homogeneous~\cite{delaparraFullApproximationDeep2020}, \texttt{mul8u\_NLX}	& \qty{29}{\percent} 		& 0.5 \\
			{}			& QoS-Nets, $o=1$, $n=3$														& \textbf{38 \%}		& \textbf{0.3} \\
			\midrule
			ResNet32	& LVRM~\cite{tasoulasWeightOrientedApproximationEnergyEfficient2020}			& \qty{18}{\percent}		& 0.5 \\
			{} 			& PNAM~\cite{spantidiPositiveNegativeApproximate2021}
			& \qty{22}{\percent}	& 1.0 \\
			{}			&  Homogeneous~\cite{delaparraFullApproximationDeep2020}, \texttt{mul8u\_NLX}	& \qty{29}{\percent} 		& \textbf{0.2} \\
			{}			& QoS-Nets, $o=1$, $n=3$														& \textbf{40 \%}		& 0.5 \\
			\bottomrule
		\end{tabulary}
		\label{tab:resnet_c10}
	\end{table}
	\begin{table}[tb]
		\caption{Comparison of power reduction and accuracy loss on CIFAR-100 for different ResNet variants and multiplier selection methods}
		\footnotesize
		\begin{tabulary}{\linewidth}{llCC}
			\toprule
			Model & Method & Power Reduction \textuparrow & Top-1 Accuracy Loss [p.p.] \textdownarrow \\
			\midrule
			ResNet20	& TPM~\cite{spantidiPerfectMatchSelecting2023}								& \qty{3}{\percent}	& 0.5 \\
			{} & PNAM~\cite{spantidiPositiveNegativeApproximate2021} &
			\qty{20}{\percent}	& 0.5 \\
			{}			& QoS-Nets, $o=1$, $n=3$															& \textbf{21 \%}	& \textbf{0.0} \\
			\midrule
			ResNet32	& TPM~\cite{spantidiPerfectMatchSelecting2023}								& \qty{3}{\percent}	& 0.5 \\
			{} & PNAM~\cite{spantidiPositiveNegativeApproximate2021} &
			\qty{22}{\percent}	& 0.5 \\
			{}			& QoS-Nets, $o=1$, $n=3$															& \textbf{24 \%}	& \textbf{-0.2} \\
			\bottomrule
		\end{tabulary}
		\label{tab:resnet_c100}
	\end{table}
	\label{sec:resnet_cifar}
	The results for CIFAR10 in~\Cref{tab:resnet_c10} and CIFAR100 in~\Cref{tab:resnet_c100} show that the combination of retraining and heterogeneous \ac{am} assignment is consistently on par with or improves upon existing algorithms. Only the retrained homogeneous solution provides a comparable ratio of performance and reduction in power consumption in some of the test cases. Homogeneous solutions, however, only provide a comparatively sparse Pareto front, making it hard to find comparable energy/accuracy trade-offs in some cases. The large performance boost through retraining underlines the importance of optimizing \ac{nn} parameters to match the approximate computation error.
	
\subsection{Memory-efficient QoS-Nets}
\label{sec:memory_efficient_qos}
We assess the performance of the QoS-Nets methodology on an application that is more complex than ResNet8/CIFAR. As a dataset, we choose TinyImageNet, a downscaled version of the full ImageNet with 200 classes (instead of 1000), and with images of 64\texttimes64 px resolution (down from 224\texttimes224 px). MobileNetV2~\cite{sandlerMobileNetV2InvertedResiduals2018}, a modern \ac{cnn} that is commonly used for constrained image recognition tasks, is used as the model architecture. To account for the lower resolution, we reduce the stride of the initial convolution to one, down from two.

\begin{figure}[tb]
	\centering
	\includegraphics[width=\linewidth]{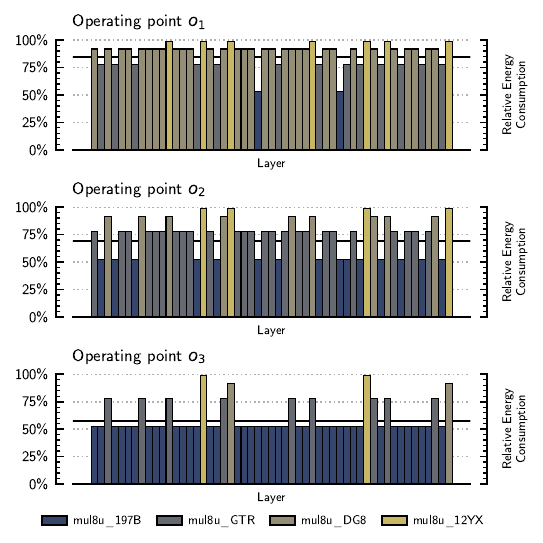}
	\caption{Multiplier choice and assignment result for the QoS-Nets configuration discussed in~\Cref{sec:memory_efficient_qos}. Horizontal line marks the network's combined relative power consumption for multiplications at each operating point.}
	\label{fig:mnet_assignment}
\end{figure}
A baseline model is generated with 30 epochs of training and 9 epochs of \ac{qat}. Layer sensitivity is determined through a gradient search with $\lambda=0.1$, $\sigma_{\text{max}}=0.05$ and $\sigma_{\text{initial}}=0.001$ for 5 epochs. Layer accuracy preferences are rescaled with $\mathcal{S} = \left\lbrace 0.1, 0.3, 1.0 \right\rbrace$ for $o=3$ operating points and the search space is clustered into $n=4$ \ac{am} instances. The resulting assignments for the 53 target layers in MobileNetV2 are shown in~\Cref{fig:mnet_assignment}. The plots show how, for each layer, the \ac{am} instance changes between the different operating points. The search picks \ac{am} instances with between \qty{47}{\percent} and \qty{1.3}{\percent} power reduction. The evaluation does not consider dynamic energy consumption required for switching between operating points since it is strongly dependent on a specific hardware implementation and small, compared to a full inference pass. Because the switching is only meant to happen \emph{between} inference passes, the reported accuracy losses are deterministic. The resulting configurations are retrained for two epochs with a learning rate of $2 \cdot 10^{-3}$ and $2 \cdot 10^{-4}$.

\begin{table}[tb]
	\caption{Relative power consumption for multiplications (underlined) and Top-5 Accuracy Loss (italic) for MobileNetV2/TinyImageNet across $o=3$ operating points for different multiplier selection and retraining methods}
	\scriptsize
	\begin{tabulary}{\linewidth}{@{\extracolsep{4pt}}LRRRRR}
		\toprule
		\multirow{2.2}{*}{\footnotesize Method} & \multicolumn{3}{c}{\makecell{\underline{Relative Power Consumption} \textdownarrow \\\textit{Top-5 Acc. Loss [p.p.]} \textdownarrow}} & \multirow{2.2}{*}{\footnotesize \#AMs \textdownarrow} & \multirow{2.2}{*}{\footnotesize Params. \textdownarrow}\\ \cline{2-4}
		{}	& $o_1$ & $o_2$ & $o_3$ & {} & {} \\[2pt]
		\midrule
		Baseline Model & \multicolumn{3}{c}{\rule[2pt]{4em}{0.3pt} \makecell{\underline{100\%}\\\textit{0.0} (79.21\%)*} \rule[2pt]{4em}{0.3pt}} &  --- & 2.48M \\ \midrule
		\vc{Homogeneous} & \underline{84.14\%} \textit{0.85}   & \underline{70.59\%} \textit{0.51}  & \underline{60.61\%} \textit{15.86} & \vc{3} & \vc{7.44M}\\ \midrule
		Gradient Search~\cite{trommerCombiningGradientsProbabilities2022} & \underline{83.68\%} \textit{0.08}   & \underline{70.48\%} \textit{0.47}  & \underline{55.86\%} \textit{2.02} & \vc{16} & \vc{7.44M}\\ \midrule
		QoS-Nets & \underline{84.73\%} & \underline{69.38\%} & \underline{57.19\%} & \multirow{4}{*}{4} & {}\\
		$\rightarrow$ w/o retraining & \textit{30.01} & \textit{76.79} & \textit{76.71} & {} & 2.48M\\
		$\rightarrow$ full retraining & \textit{0.10} & \textit{0.52} & \textit{1.65} & {} & 7.44M\\
		$\rightarrow$ BN Tuning & \textit{0.30} & \textit{0.71} & \textit{2.33} &  {} & 2.54M\\
		\bottomrule
	\end{tabulary}\\[3pt]
	\raggedleft
	{\scriptsize * Baseline Top-5 Accuracy of model w. 8-Bit Quantization, w/o approximate computation}
	\label{tab:mnetv2}
\end{table}
The performance and parameter counts for different retraining strategies are compared in~\Cref{tab:mnetv2}. No model can achieve useful accuracy without some form of retraining, with operating points $o_2$ and $o_3$ showing an accuracy degradation where the performance is equivalent to random guessing. The proposed fine-tuning of only the BatchNormalization is able to recover most of the model's accuracy, with full retraining only giving an average 0.35 percentage points higher accuracy at the cost of a 200\% parameter overhead, while our proposed fine-tuning scheme only adds 2.75\% additional parameters. This finding suggests that \emph{only} retraining a layer's biases or batch normalization during the fine-tuning stage while leaving the weights fixed is sufficient to recover nearly all of the model's accuracy.

To compare QoS-Nets to the unconstrained Gradient Search algorithm, the multiplier matching method that is described in the original work on the rescaled operating points, without any clustering in-between. Because of the lack of constraints, the method can minimize the \ac{am} power consumption for each layer individually, leading to lower power consumption for each operating point, at the price of a slightly reduced accuracy of 0.1 percentage points on average. Compared to our proposed fine-tuning scheme, Gradient Search with full retraining for each model delivers 0.26 percentage points higher accuracy, albeit at an impractical output space of 16 \ac{am} instances. Compared to Gradient Search, the constrained QoS-Nets search algorithm achieves a very similar performance at a fraction of the parameters and with only 4 different \ac{am} instances. We also include a comparison with three homogeneous solutions after individual retraining of all parameters. \texttt{mul8u\_8U3}, \texttt{mul8u\_NLX} and \texttt{mul8u\_1DMU} were chosen because they provide a similar power consumption to that of the three operating points determined by QoS-Nets. Particularly for $o_3$ with the lowest power consumption, the heterogeneous solutions provide much higher accuracy at lower power consumption because of their ability to adjust the provided accuracy, depending on each layer's sensitivity.

\section{Conclusion and Outlook}
This work proposed the concept of \ac{qos} scaling for systems that can dynamically adjust operator accuracy at runtime. From a large set of approximate operators, a combination of gradient-based search for layer robustness, followed by k-Means Clustering of the results picks a subset of the most useful \acp{am} and provides an assignment of each layer at each operating point to one of those \ac{am} instances.

We achieve parameter-efficient retraining by optimizing only the normalization parameters between layers while sharing the weights between all operating points. Our experimental evaluation shows that optimizing a small set of important parameters is sufficient to provide nearly the same accuracy as retraining of the entire model.

The primary focus of this work was hardware selection and parameters optimization for reconfigurable platforms. A study of this method on actual hardware remains as a topic for future research. We assume that applying the proposed method to other operator-level approximations~\cite{blalockMultiplyingMatricesMultiplying2021} is possible as long as their error is deterministic, and they provide a large search space of accuracy/resource consumption trade-off points.

We hope that \ac{qos} scaling can help increase the flexibility of \ac{nn} accelerator platforms in order to adapt to changing environmental conditions with graceful degradation of performance, rather than a binary failure mode.
\bibliographystyle{ACM-Reference-Format}
\bibliography{bibliography}	
\end{document}